\documentclass[10pt,twocolumn,letterpaper]{article}

\usepackage[pagenumbers]{iccv} 

\usepackage{times}
\usepackage{epsfig}
\usepackage{graphicx}
\usepackage{amsmath}
\usepackage{amssymb}
\usepackage{adjustbox}
\usepackage{booktabs} 
\usepackage{multirow}
\usepackage{natbib} 
\usepackage{colortbl} 

\usepackage{float}
\usepackage{afterpage}

\definecolor{iccvblue}{rgb}{0.21,0.49,0.74}
\usepackage[pagebackref,breaklinks,colorlinks,allcolors=iccvblue]{hyperref}

\title{ETVA: Evaluation of Text-to-Video Alignment \\via Fine-grained Question Generation and Answering}

\author{
    Kaisi Guan $^{1,2}$\thanks{The work was done during an internship at Apple.\quad Email: {\tt guankaisi@ruc.edu.cn}} \quad
    Zhengfeng Lai$^{2}$\quad
    Yuchong Sun$^{1}$\quad
    Peng Zhang$^{2}$\quad \\
    Wei Liu$^{2}$\quad 
    Kieran Liu$^{2}$\quad 
    Meng Cao$^{2}$\quad
    Ruihua Song$^{1}$\textsuperscript{$\dag$\thanks{\textsuperscript{$\dag$}Corresponding authors.}}\vspace{0.2em}\\
    $^1$ Renmin University of China\quad$^2$Apple \\
    \url{https://eftv-eval.github.io/etva-eval}
}

\begin{document}
\maketitle

\begin{abstract}
Precisely evaluating semantic alignment between text prompts and generated videos remains a challenge in Text-to-Video (T2V) Generation. Existing text-to-video alignment metrics like CLIPScore only generate coarse-grained scores without fine-grained alignment details, failing to align with human preference.
To address this limitation, we propose \textbf{ETVA}, a novel \textbf{E}valuation method of \textbf{T}ext-to-\textbf{V}ideo \textbf{A}lignment via fine-grained question generation and answering.
First, a multi-agent system parses prompts into semantic scene graphs to generate atomic questions.
Then we design a knowledge-augmented multi-stage reasoning framework for question answering, where an auxiliary LLM first retrieves relevant common-sense knowledge (e.g., physical laws), and then video LLM answer the generated questions through a multi-stage reasoning mechanism.
Extensive experiments demonstrate that ETVA achieves a Spearman's correlation coefficient of 58.47, showing much higher correlation with human judgment than existing metrics which attain only 31.0.
We also construct a comprehensive benchmark specifically designed for text-to-video alignment evaluation, featuring 2k diverse prompts and 12k atomic questions spanning 10 categories.
Through a systematic evaluation of 15 existing text-to-video models, we identify their key capabilities and limitations, paving the way for next-generation T2V generation. All codes and datasets will be publicly available soon.
\end{abstract}
   

\section{Introduction}
Text-to-Video (T2V) generation models ~\cite{kong2024hunyuanvideo,ma2024latte,opensora,lin2024opensora-plan,fan2025vchitect,yang2024cogvideox, kling, sora2024, pika} have shown remarkable advancements in recent years, showing particular potential in applications such as video content editing~\cite{molad2023dreamix,sun2024videoedit}, movie production~\cite{polyak2024moviegen,zhu2023moviefactoryautomaticmoviecreation}, and even the world simulator~\cite{sora2024,worldmodels,meng2024worldsimulatorcraftingphysical}.
\begin{figure}[t]
    \centering
    \includegraphics[width=\linewidth]{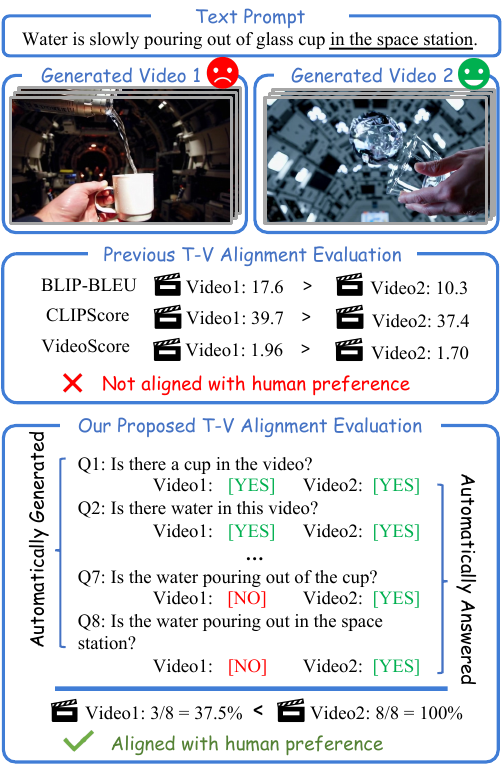}
    \caption{Illustration of how ETVA works and comparison with existing metrics.}
    \label{fig:teaser}
    \vspace{-6mm}
\end{figure}
However, current T2V generation suffers from the absence of reliable automatic metrics for text-to-video alignment. Existing text-to-video alignment metrics fall into three categories: Caption-based metrics like BLIP-BLEU~\cite{liu2024evalcrafter}, CLIP-based like CLIPScore~\cite{radford2021clip}, and MLLM-based like VideoScore~\cite{he2024videoscore}. These metrics share a common limitation: they only produce coarse-grained scores, lacking the granularity to capture fine-grained alignment details, resulting in significant deviations from human preference.  For instance, as illustrated in Figure~\ref{fig:teaser}, human annotator recognizes Video 2 as superior in depicting microgravity physics (implied by ``space station"), while existing metrics systematically favor Video 1.
Some works in text-to-image alignment~\cite{hu2023tifa,JaeminCho2024dsg,chen2024interleaveddsg,sun2023dreamsync} solve this problem by constructing a Question Generation (QG) and Question Answering (QA) pipeline with the help of Large Language Models (LLMs)~\cite{openai2024gpt4o,qwen2.5,touvron2023llama,touvron2023llama2} and Multimodal Large Language Models (MLLMs)~\cite{liu2023llava,liu2024llavanext,liu2023improvedllava}. However, these initial attempts still have not been transferred to the text-to-video alignment evaluation yet due to the following challenges:


\textit{C1.} Appropriate Questions Challenge. Video text prompts are often lengthy and contain multiple elements, making question generation complicated. Vanilla in-context learning methods, like in TIFA~\cite{hu2023tifa}, tend to generate overly intricate questions but they pose challenges for video LLMs to give correct answers. For instance, in Figure~\ref{fig:teaser}, a question like ``Is the water pouring out of a glass cup and is it in the space station?” requires simultaneous comprehension of multiple factors, including actions, objects, and environment, which poses significant challenges for Video LLMs. 

\textit{C2.} Video LLMs Hallucination Challenge. Video LLMs~\cite{wang2024qwen2-vl,openai2024gpt4o,lin2023videollava,zhang2024llavavideo} exhibit more severe hallucination issues than their image-based counterparts. Some questions require additional commonsense knowledge that is naturally accessible to human brain but unavailable to Video LLMs. As shown in Figure~\ref{fig:teaser}, when assessing whether water is pouring in a space station, humans intuitively recognize the microgravity environment, whereas Video LLMs lack this innate understanding. Moreover, while the human brain engages in deep observation and reasoning to analyze video content, Video LLMs generate responses directly, bypassing the intermediate cognitive processes.
To address these challenges, we propose \textbf{ETVA}, a novel \textbf{E}valuation method for \textbf{T}ext-to-\textbf{V}ideo \textbf{A}lignment
that contains a multi-agent framework for atomic question generation (C1) and a knowledge-augmented multi-stage reasoning framework to emulate human-like reasoning in question answering (C2). 
The QG part of ETVA consists of three collaborative agents: \textit{Elements extractor} decomposes the text prompt into three categories of elements: entities (key objects), attributes (descriptive properties), and relations (spatial-temporal interactions). Then \textit{Graph builder} structures these elements into a scene graph that covers all semantic requirements for the target video. Finally, \textit{Graph traverser} traverses each node of the graph to generate questions.
In the QA phase, ETVA adopts a brain-inspired cognitive architecture to mitigate hallucination through knowledge-augmented multi-stage reasoning. An auxiliary LLM provides commonsense knowledge augmentation (e.g. microgravity in the space station) like the brain memory area and then together with video content and questions, a multi-stage reasoning process lets video LLMs think step by step to answer the question. 
Extensive experiments demonstrate that ETVA achieves a Spearman’s correlation of 57.8, outperforming existing state-of-the-art metric VideoScore~\cite{he2024videoscore} by 27.5. Furthermore, its well-designed QG and QA parts respectively contribute to performance improvements of 14.67\% and 26.2\%.

Based on ETVA, we further construct ETVA-Bench, a comprehensive benchmark for text-to-video alignment evaluation. ETVA-Bench comprises 2K diverse prompts and 12K generated questions categorized into 10 distinct types. We evaluate 10 open-source and 5 closed-source T2V models, providing a systematic study with in-depth analysis. Our findings reveal that these models still struggle in some areas such as camera movements or physics process.

\noindent Our contributions can be summarized as follows:
\begin{itemize}
    \item  We propose ETVA, a novel text-video alignment evaluation method with a multi-agent framework for question generation and a knowledge-augmented multi-stage reasoning framework for question answering.  
    \item We build ETVABench, a benchmark dedicated to text-video alignment evaluation, providing a systematic comparison and analysis of existing 15 T2V models.
   \item Extensive experiments on both question generation and question answering validate the effectiveness of our approach, demonstrating superior human alignment compared to existing evaluation metrics.
\end{itemize}

\section{Related Work}
\paragraph{Text-to-Video Generation}
\begin{figure*}[!ht]
    \centering
    \includegraphics[width=\linewidth]{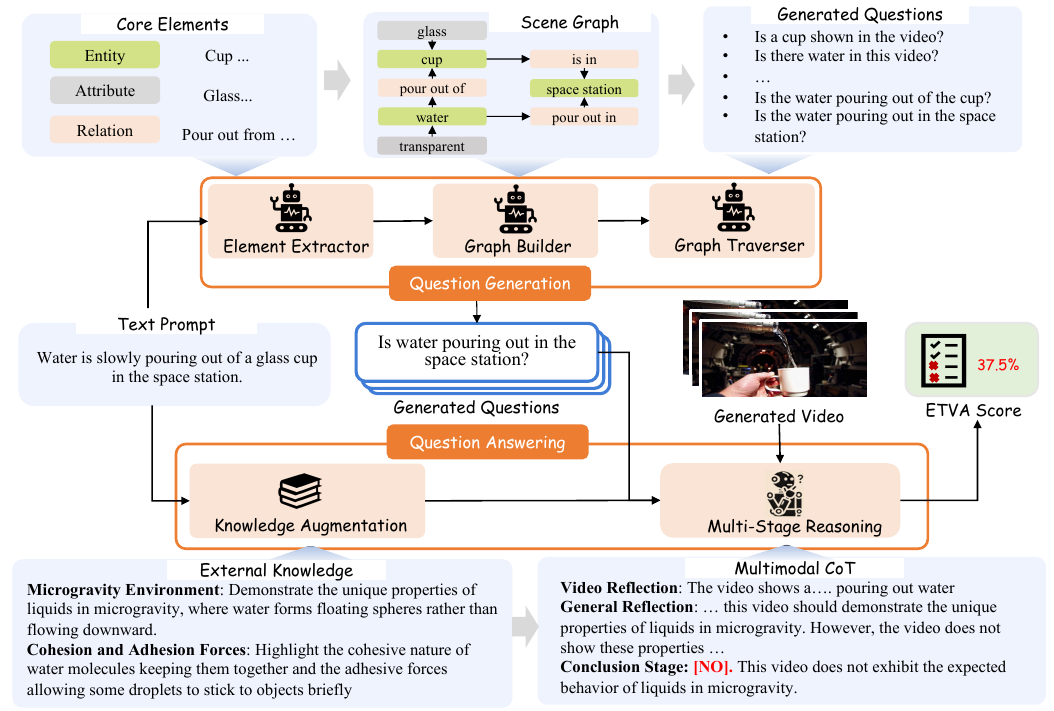}
    \vspace{-6mm}
    \caption{Overall pipeline of ETVA. ETVA contains a multi-agent framework for generating atomic questions and a knowledge-augmented multi-stage reasoning framework for question answering.}
    \label{fig:main}
\end{figure*}
Text-to-Video (T2V) models aim to generate videos from textual descriptions. There are three categories of architecture: (1) Some early works~\cite{clark2019dvd-gan,yan2021videogpt} employ Generative Adversarial Networks (GAN~\cite{goodfellow2014gan}) or Vector Quantisation Variational Autoencoders (VQ-VAE~\cite{oord2018vqvae}), yet these approaches often yield low-quality outputs with limited generalizability.
(2) UNet-based Diffusion models, such as Stable Video Diffusion~\cite{blattmann2023stablevideodiffusionscaling}, ModelScope~\cite{wang2023modelscope}, and Cogvideo~\cite{hong2022cogvideo} leverage 3D-UNet~\cite{3dunet} architectures to effectively model both spatial and temporal information.
(3) Diffusion Transformers (DiT)~\cite{Peebles2022DiT}. DiT represents the state-of-the-art (SOTA) backbone for T2V models. A range of popular open-source models, including HunyuanVideo~\cite{kong2024hunyuanvideo}, CogVideox~\cite{yang2024cogvideox}, and OpenSora~\cite{opensora}, as well as commercial models such as Pika~\cite{pika}, Vidu~\cite{vidu}, Sora~\cite{sora2024}, MetaMovieGen~\cite{polyak2024moviegen} and Kling~\cite{kling}, have adopted DiT as their core architecture. In this paper, we focus on the DiT-based T2V models.

\paragraph{Automatic Text-to-Video Alignment Metrics }
Existing work evaluates text-to-video alignment using three methods: (1) Caption-based metrics, where caption models (e.g. BLIP~\cite{li2022blip,li2023blip2}) generate video caption followed by calculating text-text similarity between caption and prompt using metrics like BLEU~\cite{bleu}; (2) CLIP-based metrics~\cite{radford2021clip,wang2024viclip,li2024unmaskedteachertrainingefficientvideo}, extracting semantic embeddings from text and video and calculating similarity, like ViCLIP~\cite{wang2024viclip} in VBench~\cite{huang2023vbench}. (3) MLLM-based metrics, which fine-tune~\cite{he2024videoscore} Multimodal Language Models (MLLM) on a human-annotated dataset to be a reward model for scoring, such as VideoScore~\cite{he2024videoscore}. However, these methods often lack fine-grained detail and do not correlate well with human judgments.
Some text-to-image alignment evaluation works~\cite{hu2023tifa,JaeminCho2024dsg,sun2023dreamsync,chen2024interleaveddsg} use automatic QG \& QA framework, achieving better alignment with human preference. However, due to complexities of text-to-video evaluation~\cite{liu2024surveyaigeneratedvideoevaluation,liao2024devil}, these methods cannot transfer directly to text-to-video alignment.

\paragraph{Benchmarks for T2V Generation}
Existing T2V generation benchmarks can be categorized into two groups: (1) General benchmarks evaluate the overall performance of the T2V model, including quality, consistency, and aesthetics (e.g., VBench~\cite{huang2023vbench}, EvalCrafter~\cite{liu2024evalcrafter}, MetaMovieGen~\cite{polyak2024moviegen}, FETV~\cite{liu2023fetv}), however, they often overlook fine-grained details. (2) Specific Benchmarks focus on particular aspects of T2V generation, such as GAIA~\cite{chen2024gaia} to evaluate human actions, T2V-ComBench~\cite{sun2024t2vcompbench} for multi-object composition, VideoPhy ~\cite{bansal2024videophy} \& PhyGenBench~\cite{meng2024phygenbench} for physics phenomenon, and ChronoMagic-Bench~\cite{yuan2024chronomagic} for time-lapse video.
In this work, we build a specific benchmark for evaluating the text-to-video alignment of T2V models.

\section{Our Proposed ETVA Approach}
\label{section3}
The main idea of ETVA is to simulate human annotation process by using LLMs and Video LLMs to automatically generate and answer questions. As illustrated in Figure \ref{fig:main}, the framework operates in two stages: (1) Question Generation:  A multi-agent system generates detailed questions using scene graphs (Section \ref{question_generation}). (2) Question Answering:  An auxiliary LLM first recalls common sense knowledge, then Video LLM analyzes videos and reasons through answers (like human thinking) (Section \ref{question_answering}).


\subsection{Problem Formulation}
Given a text prompt $T$ and a generated video $V$, our ETVA approach computes an alignment score $S \in [0,1]$ through the mapping:
\begin{equation}
    S = \text{ETVA}(T, V)
\end{equation}

\noindent Section \ref{question_generation} produces a set of binary verification questions $\mathcal{Q} = \{Q_1, \dots, Q_n\}$ derived from text prompt $T$, where each $Q_i \in \mathcal{Q}$ admits a yes/no response:
\begin{equation}
    \mathcal{Q} = G(T)
\end{equation}

\noindent Section \ref{question_answering} resolves these questions using three information sources: original text $T$, video content $V$, and generated questions $\mathcal{Q}$. First, common-sense knowledge $K$ is extracted through language model inference:
\begin{equation}
    K = \text{LLM}(T)
\end{equation}

\noindent Each binary score $S_i \in \{0,1\}$ is then answered by question $Q_i$ with knowledge $K$ and video content $V$:
\begin{equation}
    S_i = A(Q_i, K, V)
\end{equation}

\noindent The final alignment score is calculated based on results across all questions:
\begin{equation}
    S = \frac{1}{n}\sum_{i=1}^{n} S_i
\end{equation}

\subsection{Multi-Agent Question Generation}
\label{question_generation}
Generating video-relevant questions from text prompts presents a fundamental challenge. While conventional approaches employ LLM in-context learning with predefined question categories, they often suffer from semantic redundancy, incomplete coverage, and unanswerable outputs that degrade evaluation quality. Our solution borrows the idea from scene graph for getting atomic elements~\cite{krishna2017visualscenegraph,johnson2018scenegraph,teng2025atom}  to structure the question generation process. As illustrated in Figure~\ref{fig:main}, our framework implements a multi-agent framework containing three agents:
\begin{itemize}
    \item \textit{Element Extractor} identifies core textual elements
    \item \textit{Graph Builder} constructs semantic relationships
    \item \textit{Graph Traverser} generates comprehensive questions
\end{itemize}
This collaborative architecture ensures fine-grained questioning with complete semantic coverage through systematic graph exploration.
\subsubsection{Element Extractor}  
The \textit{Element Extractor} identifies key elements in text prompts through three fundamental types: entities (specific objects like ``cup" and ``space station"), attributes (descriptive features like ``glass material" and ``transparent" for water), and relationships (action/positional links like ``pouring out of" and ``contained within"). For instance, in Figure~\ref{fig:main} with the prompt "Water is slowly pouring out of a glass cup in the space station," it detects entities (water, cup, space station), material attributes (glass, transparent), and spatial relationships (pouring-from, contain in). This structured decomposition enables systematic question generation by parsing key elements from textual descriptions.
\subsubsection{Graph Builder}
The \textit{Graph Builder} constructs hierarchical scene graphs where nodes represent elements identified by the \textit{Element Extractor}. Adjacent nodes belong to distinct categories, with entity nodes (e.g., objects) serving as central anchors - all relation nodes (actions/spatial links) and attribute nodes (descriptive features) must connect to at least one entity node. As shown in Figure~\ref{fig:main}, attribute nodes only emit outgoing edges (e.g., ``glass → material''), while relation nodes maintain bidirectional connections between entities (e.g., ``cup ← pouring → water''). This structure ensures semantic coherence by enforcing category exclusivity between neighboring nodes and preventing redundant connections through entity-centric topology. 
\subsubsection{Graph Traverser} The \textit{Graph Traverser} systematically explores scene graphs to generate yes/no questions through a three-step process: It first handles main objects (entities), then their attributes (descriptive details), and finally checks connections (relations) between objects – but only after both objects in the connection have been processed. As shown in Figure~\ref{fig:main}, this ordered approach ensures natural question flow by first confirming object properties (e.g., ``Is the cup made of glass?") before examining interactions (e.g., ``Is water pouring from the cup?"). The system automatically delays relationship questions until both connected objects have their attributes verified, using this smart order to avoid confusing questions and maintain clear context. This dependency-checking method guarantees that every question builds logically on previously confirmed information.

\subsection{Multi-modal Question Answering}
In this part, we aim to leverage video LLM to answer generated questions. Video-related questions often require both additional contextual knowledge and thorough comparative analysis for accurate answers. Direct video LLM responses often suffer from severe hallucination problems. Inspired by human annotation processes – such as evaluating the prompt ``Water slowly pours from a glass cup in the space station" – where annotators first recall relevant common sense (e.g., liquid behavior in microgravity) before synthesizing knowledge with visual observations, we implement an automated simulation of this cognitive workflow in ETVA-Rater question answering part. As shown in Figure~\ref{fig:main}, the framework's knowledge augment component emulates human memory recall, while its multistage reasoning mechanism enables comprehensive analysis.

\label{question_answering}
\subsubsection{Knowledge Augmentation with auxiliary LLM}
In this stage, we leverage an auxiliary LLM to provide common-sense knowledge for video understanding. Large Language Models (LLMs) undergo extensive pretraining, allowing them to internalize knowledge similar to the human brain. By carefully designing prompts, we can guide the LLM to recall relevant knowledge for video comprehension. Here, we employ Qwen2.5-72B-Instruct~\cite{qwen2.5}. The LLM generates detailed video descriptions to enhance video LLM reasoning. For example, given the prompt ``Water is slowly pouring from a glass cup in the space station,” it identifies two key physical principles:
(1) In microgravity, liquids form floating spheres rather than falling streams.
(2) Due to the cohesion force, water should condense together.
This explicit knowledge defines expected physical behaviors, facilitating systematic video analysis.
\label{knowledge_augment}

\subsubsection{Multi-Stage Reasoning} This phase leverages the video LLM's multi-modal chain-of-thought (CoT) capabilities through three progressive steps to fully analyse the video and question. 
\begin{itemize}
    \item Video Understanding Stage: Video LLM autonomously extracts visual patterns by generating frame-by-frame descriptions without text input.
    \item General Reflection Stage: Video LLM combines these observations with the question context and common-sense knowledge for cross-modal analysis, identifying key evidence through iterative verification.
    \item Conclusion Stage: Video LLM delivers a conclusive Yes/No answer supported by explicit visual-linguistic alignment checks, ensuring decisions remain grounded in both the video content and logical reasoning.
\end{itemize}
\section{Our Proposed ETVABench}
In this section, we introduce ETVABench, an automatic Text-to-Video generation alignment benchmark based on ETVA. We collect prompts from several open-source benchmarks (Section \ref{collection}) and use a question-driven classification method to decompose these prompts to 10 distinct categories (Section \ref{classification}). 

\subsection{Our Prompts Collection and Sources}
\label{collection}
We first build ETVABench-2k which collects 2k text prompts from diverse open-source benchmarks to ensure comprehensive coverage of text-to-video alignment challenges. Specifically, we sample 880 general prompts from established T2V evaluation frameworks (VBench~\cite{huang2023vbench}, EvalCrafter~\cite{liu2024evalcrafter}, VideoGen-Eval~\cite{zeng2024videogeneval}), 300 compositional prompts from T2V-ComBench~\cite{sun2024t2vcompbench}, and 510 human motion descriptions from GAIA~\cite{chen2024gaia}. The dataset further incorporates 160 physics-aware prompts from VideoPhy~\cite{bansal2024videophy} and PhyGenBench~\cite{meng2024phygenbench}, 100 temporal coherence queries from ChronoMagic-Bench~\cite{yuan2024chronomagic}, supplemented by 50 GPT-4-o~\cite{openai2024gpt4o} generated prompts targeting hyper-realistic scenarios. The multiple sources ensures both breadth and depth in evaluating cross-modal alignment capabilities. Since evaluating closed-source models via web access is resource-intensive, we create ETVABench-105 by strategically sampling 105 prompts from the original 2,000-prompt ETVABench-2k. This compact subset preserves the full benchmark's question type distribution while enabling cost-effective testing, maintaining evaluation validity through representative prompt selection, as shown in Appendix.

\begin{table*}[t]
    \centering
    \resizebox{\linewidth}{!}{
    \begin{tabular}{lcccccccccc|c}
    \toprule
     \textbf{Metric} & \textbf{Existence} & \textbf{Action} & \textbf{Material} & \textbf{Spatial} & \textbf{Number} & \textbf{Shape} & \textbf{Color} & \textbf{Camera} & \textbf{Physics}  & \textbf{Other} & \textbf{Overall} \\
     \midrule
      BLIP-ROUGE~\cite{liu2024evalcrafter} &6.0/8.4 &6.5/8.9 & 3.0/5.3 & 9.2/12.6 & 11.0/15.5 & 9.3/14.7 & -4.7/-5.2 & 3.3/4.9 & 5.9/8.8 & 5.3/7.1 & 6.3/8.8 \\
      BLIP-BLEU~\cite{liu2024evalcrafter} & 9.3/12.9 & 8.4/11.6 & 6.2/10.1 & 11.2/15.4 & 10.0/13.6 & 9.2/12.7 & 6.6/10.5 & 10.3/14.7 & 10.1/14.4 & 8.6/11.4 & 8.5/12.1 \\
     CLIPScore~\cite{radford2021clip} & 10.2/13.7 & 10.6/13.6 & 9.9/12.8 & 12.2/16.1 & 10.9/14.8 & 14.6/20.8 & 11.8/18.8 & 12.9/15.2 & 9.7/14.2 & 10.2/12.9 & 10.3/13.8 \\
     UMTScore~\cite{li2024unmaskedteachertrainingefficientvideo} & 17.9/24.0 & 14.3/19.0 & 25.4/34.4 & 21.6/28.2 & 9.1/13.5 & 24.4/32.1 & 22.5/31.1 & 22.2/29.0 & 18.4/23.2 & 15.5/20.6 & 17.6/23.5 \\
     ViCLIPScore~\cite{wang2024viclip} & 20.2/27.1 & 17.9/24.2 & 16.2/20.4 & 20.6/27.2 & 16.9/22.4 & 29.4/38.4 & 13.8/13.5 & 23.6/31.8 & 19.8/26.3 & 17.0/22.7 & 19.4/25.9 \\
      VideoScore~\cite{he2024videoscore} & 23.2/30.6 & 22.7/30.2 & 29.9/37.3  & 24.8/31.7 & 26.6/35.9 & 28.1/35.7 & 11.7/16.2 & 19.2/26.3 & 20.3/23.9 & 23.9/31.6 & 23.7/31.0 \\
      \midrule
      \textbf{ETVA} & \textbf{47.7/57.4} & \textbf{38.3/46.6} & \textbf{55.5/66.1} & \textbf{56.0/66.8} & \textbf{44.0/53.9} & \textbf{64.1/75.1} & \textbf{31.5/39.7} & \textbf{35.5/44.2} & \textbf{50.6/60.4} & \textbf{49.0/59.2} & \textbf{47.2/58.5} \\
      \bottomrule
    \end{tabular}%
    }
    \vspace{-2mm}
    \caption{Correlations between each evaluation metric and human judgment on text alignment, measured by Kendall’s $\tau$ (left) and Spearman’s $\rho$ (right). The same category denotes groups of prompts that produce the same evaluation questions.}
    \label{tab:correlation}
    \vspace{-2mm}
\end{table*}
\subsection{Prompt Classification based on Question Labels}
\label{classification}
Existing text-to-video evaluation benchmarks typically employ coarse-grained prompt categorization, which proves inadequate for complex prompts containing multiple semantic elements. To address this limitation, we propose an atomic question-based taxonomy derived from scene graph node traversals in Section~\ref{question_generation}. Our classification framework organizes prompts into 10 distinct categories (\texttt{existence, action, material, spatial, number, shape, color, camera, physics, other}) based on their constituent atomic questions. Prompts that generate the same questions are grouped under the same category. ETVABench-2k contains 12K questions, while ETVABench-105 includes 0.6K. Further details, including the specific number of prompts in each category, are provided in the Appendix.



\section{Experiment}
\begin{table}[tp]
    \centering
    \setlength\tabcolsep{6.0pt}
    \resizebox{\linewidth}{!}{
        \begin{tabular}{l|cc}
        \toprule
        \textbf{Settings} & \textbf{Kendall's $\tau$} & \textbf{Spearsman's $\rho$} \\
        \midrule
        \textbf{Multi-agent QG} & \textbf{47.16} \textbf{($\Delta$ +12.12)} & \textbf{58.47}\textbf{($\Delta$ +15.60)} \\
       Vannila QG &35.04 & 42.87\\
        \bottomrule
        \end{tabular}
    }
    \vspace{-2mm\textbf{}}
    \caption{Ablation Study of ETVA Question Generation. Both multi-agent and vanilla QG are based on the knowledge-augmented multi-stage reasoning framework.}
    \vspace{-2mm}
    \label{tab:question_gen}
\end{table}
In this section, we first present the experimental setup (Section \ref{setting}). Then we conduct extensive experiments to validate ETVA's superior correlation with human judgment compared to existing metrics and ablation study of each part (Section \ref{human_correlation}).  Based on ETVABench, we conduct a comprehensive evaluation of text-video alignment across 15 existing T2V models and give an analysis of nowadays models limitation (Section \ref{benchmarking}). Finally, we give some case studies of ETVA on existing T2V models (Section \ref{case_study}).

\subsection{Experiment Setup} \label{setting}
\noindent \textbf{ETVA Configuration} Our framework do not have any training procedures. For question generation, we employ Qwen2.5-72B-instruct~\cite{qwen2.5} as the large language model (LLM). For question answering, we use Qwen2.5-72B-instruct~\cite{qwen2.5} as the auxiliary LLM in Knowledge Augmentation part and use Qwen2-VL-72B~\cite{wang2024qwen2-vl} as the video LLM in the Multi-stage reasoning part.  \\
\noindent \textbf{Model Evaluation} We benchmark 10 open-source and 5 closed-source text-to-video generation models on ETVABench, all implementing Diffusion Transformer (DiT)~\cite{Peebles2022DiT} architectures. Evaluations were conducted using default configurations (Details are in Appendix).
\begin{itemize}
\item \textbf{Open-Source Models}: Hunyuan-Video~\cite{kong2024hunyuanvideo}, CogVideoX-5B~\cite{yang2024cogvideox}, CogVideoX-2B~\cite{yang2024cogvideox}, CogVideoX-1.5-5B~\cite{yang2024cogvideox}, OpenSora-1.2~\cite{opensora}, OpenSora-Plan-1.2~\cite{lin2024opensora-plan}, OpenSora-Plan-1.1~\cite{lin2024opensora-plan}, Mochi-1-Preview~\cite{genmo2024mochi}, Latte~\cite{ma2024latte}, and Vchitect-2.0~\cite{fan2025vchitect}
\item \textbf{Closed-Source Models}: Sora~\cite{sora2024}, Kling-1~\cite{kling}, Kling-1.5~\cite{kling}, Pika-1.5~\cite{pika}, Vidu-1.5~\cite{vidu}
\end{itemize}
\noindent \textbf{Benchmark setting} For ETVABench-105, we evaluate both open-source and closed-source models. For ETVABench-2k, we only evaluate open-source models due to access limitations that make large-scale generation with closed-source models challenging.
\subsection{Evaluation of ETVA and Existing Metrics}
\label{human_correlation}
\subsubsection{Baselines metrics}
We compare ETVA against three categories of text-to-video alignment metrics: 
\begin{itemize}
     \item \textbf{Caption-based metrics}: BLIP-2~\cite{li2023blip2} generated captions evaluated through BLEU~\cite{papineni-etal-2002-bleu} and ROUGE~\cite{lin-2004-rouge} similarity measures against original prompts, following the setting of EvalCrafter~\cite{liu2024evalcrafter}.
    \item \textbf{CLIP-based metrics}: (i) CLIPScore~\cite{radford2021clip} (ViT-B/32) also called CLIPSIM employed in works like Pixeldance~\cite{zeng2024pixeldance} and EvalCrafter~\cite{liu2024evalcrafter}; (ii) ViCLIPScore~\cite{wang2024viclip} using video-text pretrained CLIP~\cite{radford2021clip} following the implementation of VBench~\cite{huang2023vbench}; (iii) UMTScore uses CLIP model trained on MSR-VTT~\cite{chen2022msrvtt} via unmasked teacher methods~\cite{li2024unmaskedteachertrainingefficientvideo}, following the setting of FETV~\cite{liu2023fetv}.
    \item \textbf{MLLM-based metrics}: VideoScore-Qwen2-VL~\cite{he2024videoscore}, a Qwen2-VL-7B-based~\cite{wang2024qwen2-vl} model fine-tuned for scoring video including text-alignment dimension, demonstrating superior performance to GPT-4o~\cite{openai2024gpt4o}.
\end{itemize}
\begin{table}[t]
    \centering
    \resizebox{\linewidth}{!}{%
    \begin{tabular}{lccc}
    \toprule
     Setting & \textbf{Accuracy} &\textbf{Kendall's $\tau$} & \textbf{Spearsman's $\rho$}\\
     \midrule
     \rowcolor{gray!20} 
   \textbf{ETVA} &\textbf{89.27 (+25.20)} &\textbf{47.16 (+28.98)} & \textbf{58.47 (+34.63)}\\
    ~~~~w/o. KA  &67.34 & 27.34 & 35.54\\
    ~~~~w/o. VU &82.73 & 37.56 & 44.81 \\
    ~~~~w/o. CR & 68.74 & 28.73 & 38.21 \\
    ~~~~w/Only KA & 65.48 & 24.72& 33.12\\
    Direct answer & 63.07 & 18.18 & 23.84 \\
    \bottomrule
    \end{tabular}%
    }
    \vspace{-2mm}
    \caption{Ablation Study of ETVA Question Answering Part: Component Analysis (KA: Knowledge Augmentation; VU: Video Understanding; CR: Critical Reflection)}
    \label{tab:question_answer}
    \vspace{-2mm}
\end{table}
\begin{table*}[!ht]
    \centering
\begin{adjustbox}{width=\linewidth,center}
\renewcommand{\arraystretch}{1}
\setlength{\tabcolsep}{1.2mm}
\begin{tabular}{lcccccccccc|c}
\toprule  
\textbf{Model} & \textbf{Existence} & \textbf{Action} & \textbf{Material} & \textbf{Spatial} & \textbf{Number} & \textbf{Shape} & \textbf{Color} & \textbf{Camera} & \textbf{Physics} & \textbf{Other} & \textbf{Avg} \\

\midrule
\rowcolor{gray!20} \textit{Open-Source T2V Models} & & & & & & & & & & &  \\
Latte~& 0.519 & 0.504 & 0.521 & 0.444 & 0.448 & 0.588 & 0.583 & 0.105 & 0.350 & 0.364 & 0.474 \\
Opensora-plan-1.1~ &0.514 &0.460 & 0.625 & 0.611 & 0.552 & 0.588 & 0.733 & 0.316 & 0.300 & 0.470 & 0.511  \\
Opensora-plan-1.2~ &0.524 & 0.348 & 0.684 & 0.625 & 0.578 & 0.613 & 0.783 & 0.323 & 0.300 & 0.435 & 0.527\\
Opensora-1.2~ & 0.602 & 0.606 & 0.604 & 0.574 & 0.621 & 0.588 & 0.500 & 0.263 & 0.450 & 0.424 & 0.565 \\
Cogvideox-2B~ &0.630 & 0.606 & 0.625 & 0.611 & 0.828 & 0.765 & 0.583 & 0.421 & 0.450 & 0.561 & 0.615 \\
Cogvideox-5B~ & 0.644 & 0.664 & 0.542 & 0.630 & 0.724 & 0.647 & 0.583 & \underline{\textbf{0.474}} & \underline{0.500} & 0.530 & 0.620\\
Vchitect-2.0~ &  0.745 & 0.650 & 0.667 & 0.593 & 0.621 & 0.824 & 0.333 & 0.316 & 0.350 & 0.530 & 0.644\\
Cogvideox-1.5-5B~ & 0.662 & 0.737 & 0.625 & 0.630 & \underline{0.759} & 0.765 & 0.667 & 0.368 & 0.450 & 0.545 & 0.652\\
Mochi-1-preview~& \underline{0.736} & 0.657 & \underline{0.750} & 0.685 & 0.690 & 0.588 & 0.667 & 0.421 & 0.300 & 0.636 & 0.673 \\
\underline{Hunyuan-Video}~& 0.727 & \underline{0.693} & 0.646 & \underline{0.704} & 0.724 & \underline{\textbf{0.824}} & \underline{\textbf{0.833}} & 0.421 & 0.300 & \underline{0.667} & \underline{0.686} \\
\midrule
\rowcolor{gray!20} \textit{Close-Source T2V Models} & & & & & & & & & & & \\
Kling~ & 0.730 & 0.720 & 0.558 & 0.637 & 0.755 & 0.629 & 0.600 & 0.311 & 0.450 & 0.645 & 0.671 \\
Kling-1.5 & 0.754 & 0.675 & 0.766 & 0.754 & \textbf{0.775} & 0.591 & 0.693 & 0.383 & 0.500 & 0.765 & 0.707\\
Pika-1.5   & 0.801 & 0.752 & 0.729 & 0.778 & 0.724 & 0.647 & \textbf{0.833} & 0.421 & 0.450 & 0.667 & 0.738 \\
Sora & \textbf{0.815} & 0.759 & 0.729 & \textbf{0.870} & 0.690 & 0.765 & \textbf{0.833} & 0.316 & 0.550 & 0.697 & 0.757  \\
\textbf{Vidu-1.5} & 0.792 & \textbf{0.766} & \textbf{0.854} & 0.862 & 0.714 & 0.529 & 0.667 & 0.421 & \textbf{0.600} & \textbf{0.742} & \textbf{0.761}\\
\bottomrule
\end{tabular}
\end{adjustbox}
\vspace{-2mm}
\caption{ETVABench-105 evaluation results with 5 close-source T2V models and 10 open-source T2V models. A higher score indicates better performance for a dimension. \textbf{Bold} stands for the best score. \underline{Underline} indicates the best score in the open-source models.}
\vspace{-4mm}
\label{tab:main}
\end{table*}

\subsubsection{Human Annotation}

Five human annotators were employed to assess 1,575 videos generated by 15 text-to-video (T2V) models on the ETVA-105. Each annotator performed two distinct evaluation tasks: For the first task, annotators assigned fine-grained text-alignment scores using a standardized 5-point Likert scale (0-5) with 0.5-point increments, where 0 indicates complete misalignment and 5 represents perfect semantic correspondence. The second task required the annotators to respond to binary yes/no questions automatically generated by the ETVA QG part. The majority vote (most common answer) from the five annotators determined the final answer for each question. More details are in the Appendix.


\subsubsection{Comparison of 
T2V Alignment Metrics}
As shown in Table~\ref{tab:correlation}, ETVA achieves superior alignment with human judgment across all semantic categories, attaining 47.16 Kendall's $\tau$ and 58.47 Spearman's $\rho$, establishing new state-of-the-art performance. ViCLIPScore~\cite{wang2024viclip} from VBench~\cite{huang2023vbench} achieves the best correlation among CLIP-based metrics, likely benefiting from its video-text pretaining. While conventional metrics like VideoScore~\cite{he2024videoscore} show competitive performance, they share a critical limitation: their coarse-grained scoring mechanisms fail to capture fine-grained details of the video, particularly in color, camera, and physics. In contrast, our method simulates human evaluation processes by verifying atomic video-text relationships through structured question generation and knowledge-augmented reasoning. 

\begin{figure}[!ht]
    \centering
    \includegraphics[width=\linewidth]{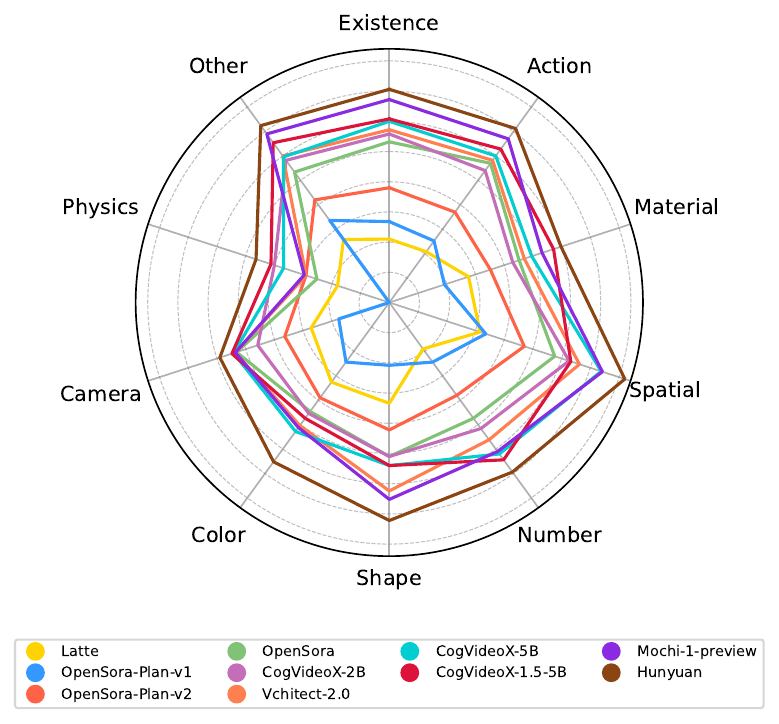}
    \caption{Evaluating results of 10 opensource T2V models in ETVABench-2k.}
    \label{fig:full}
    \vspace{-2mm}
\end{figure}
\subsubsection{Ablation Study of ETVA}
\begin{figure*}[!ht]
    \centering
    \includegraphics[width=\linewidth]{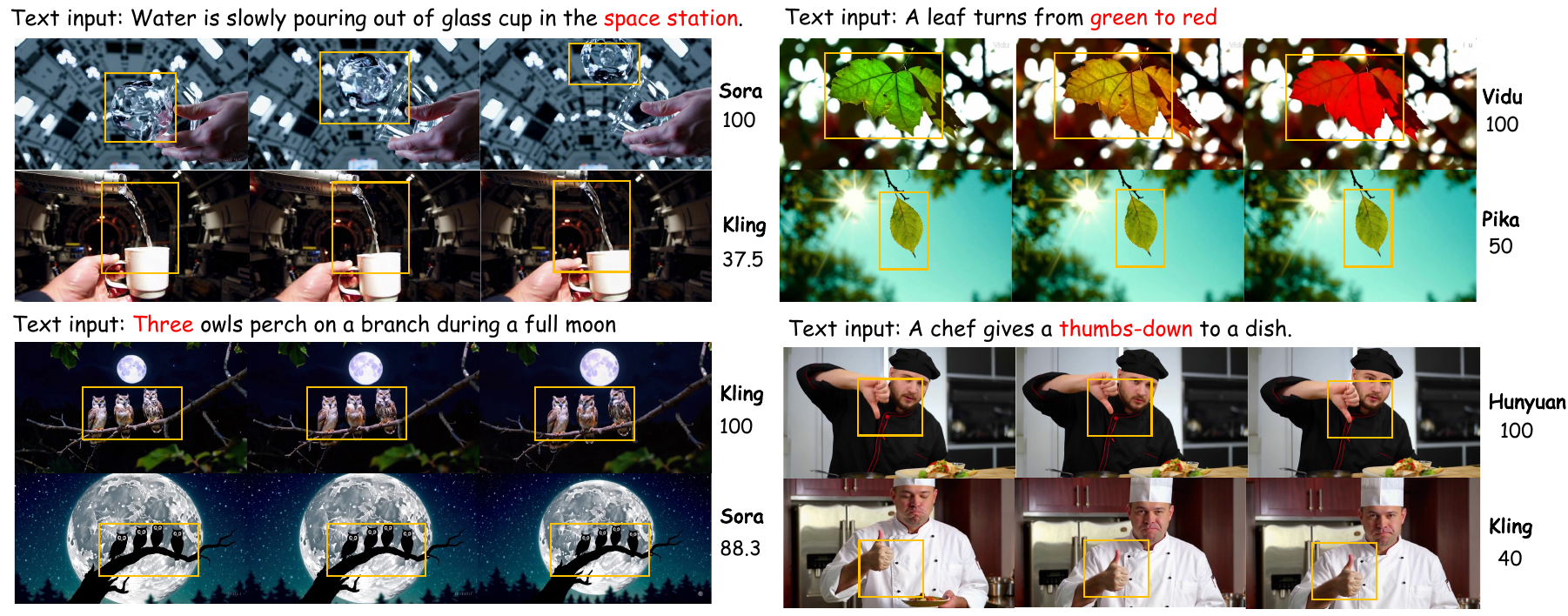}
    \caption{Four Cases of using ETVA to evaluate text-to-video alignment in T2V models, covering physics phenomenon, color transition, number accuracy and gestrual semnatic.}
    \label{fig:case}
    \vspace{-6mm}
\end{figure*}
\paragraph{Ablation of Question Generation Part} 
Vanilla question generation employs in-context learning, where the LLM generates questions directly based on prompt instructions. Our multi-agent framework improves Kendall's $\tau$ by 34.6\% and Spearman's $\rho$ by 34.1\% over vanilla in-context learning (Table~\ref{tab:question_gen}), demonstrating that structured scene graph traversal generates more discriminative questions. The performance gap stems from redundant/unanswerable questions in vanilla QG caused by incomplete semantic parsing, whereas our graph-based method ensures comprehensive coverage of entities, attributes, and relations.  
\paragraph{Ablation of Question Answering Part}
The ETVA achieves 89.27\% accuracy, surpassing direct Video LLM answers by 26.20\% (Table~\ref{tab:question_answer}). Removing the Knowledge Augment (KA) part causes the steepest performance drop of 21.93\% accuracy, highlighting its critical role in combating hallucinations. Video understanding (VU) and critical reflection (CR) contribute 6.5\% and 11.2\% accuracy gains respectively, proving that structured multimodal reasoning is irreplaceable. Notably, relying solely on KA improves accuracy by only 2.35\%, highlighting the essential role of multi-stage reasoning in effectively integrating knowledge for question answering.

\subsection{Evaluation of Existing T2V Models}
\subsubsection{Evaluation on ETVABench-105}
Table~\ref{tab:main} reveals critical insights through systematic evaluation of 10 open-source and 5 closed-source T2V models. Three key findings emerge: (1) Closed-source models dominate overall performance (Vidu-1.5: 0.761 avg), yet top open-source models like Hunyuan-Video achieve comparable capabilities in static attributes (shape: 0.824 vs Sora's 0.765); (2) All models struggle with temporal dynamics, particularly in physics (max 0.600) and camera control (max 0.474), exposing fundamental limitations in spatiotemporal reasoning; (3) Performance gaps widen for dynamic scenarios - while closed-source models excel in action (Vidu-1.5: 0.766) and spatial relations (Sora: 0.870), open-source alternatives show 18-32\% relative degradation. The results highlight the urgent need for enhanced temporal modeling and commonsense integration in T2V generation.
\subsubsection{Evaluation on ETVABench-2k}

\label{benchmarking}
Figure \ref{fig:full} presents the evaluation results of 10 open-source models on ETVABench-2k. The results show a high degree of consistency with those obtained on ETVABench-105, indicating stable performance trends across different evaluation scales. Among these models, Hunyuan-Video~\cite{kong2024hunyuanvideo} achieves the best overall performance, surpassing all other models in multiple evaluation dimensions. This suggests that Hunyuan-Video has strong text-to-video generation capabilities.
However, despite these advancements, significant challenges remain. In the Physics dimension, all models struggle to accurately simulate real-world physical interactions, often failing to maintain consistency in object motion, forces, and environmental effects. Similarly, in the Camera dimension, many models exhibit difficulties in perspective control, shot composition, and smooth transitions between frames, leading to inconsistencies in visual presentation. These limitations underscore the need for further improvements in modeling real-world dynamics and enhancing visual coherence. Addressing these issues will be crucial for advancing text-to-video generation. Detailed results and analyses are provided in the Appendix.
\subsection{Case Study}
\label{case_study}
Figure \ref{fig:case} presents several evaluation cases using ETVA to assess the text-to-video alignment of T2V models. These cases illustrate how well different models capture fine-grained semantic details, which are often overlooked by conventional evaluation metrics.
In the first case, Sora accurately depicts water pouring in a space station, correctly reflecting the effects of microgravity (Score: 100), while Kling fails to model these effects realistically, resulting in an inaccurate depiction (Score: 37.5). The second case evaluates color transformation, where Vidu successfully changes a leaf from green to red as described (Score: 100), whereas Pika produces an incomplete transition, leaving parts of the leaf unchanged (Score: 50).
The third case examines numerical accuracy. Kling correctly generates three owls as specified in the prompt (Score: 100), while Sora incorrectly produces four, demonstrating a minor but notable deviation (Score: 88.3). The final case assesses the comprehension of gestural semantics. Hunyuan faithfully executes the thumbs-down gesture as described in the text (Score: 100), while Kling misinterprets the instruction, producing an inverted hands-up gesture instead (Score: 40).

\section{Conclusion}
In this paper, we introduce ETVA, a novel evaluation method for text-to-video alignment based on fine-grained question generation and answering. ETVA addresses two key issues in existing evaluation methods: (1) Appropriate Question Challenge – We generate atomic and semantically comprehensive questions via a multi-agent question generation approach. (2) Video LLMs Hallucination Challenge – We design a knowledge-augmented, multi-stage reasoning framework for question answering, significantly reducing hallucinations.
Based on ETVA, we construct ETVABench, a benchmark with 2K diverse prompts and 12K generated questions, categorized into ten distinct types, to evaluate current T2V models. Extensive experiments show that ETVA achieves a much higher correlation with human preferences than existing metrics. It also helps identify the key limitations of T2V models, paving the way for the next generation of text-to-video models.

\newpage

{
    \small
    \bibliographystyle{ieeenat_fullname}
    \bibliography{main}
}

\clearpage
\setcounter{page}{1}
\maketitlesupplementary
\begin{figure*}[!ht]
    \centering
    \includegraphics[width=\linewidth]{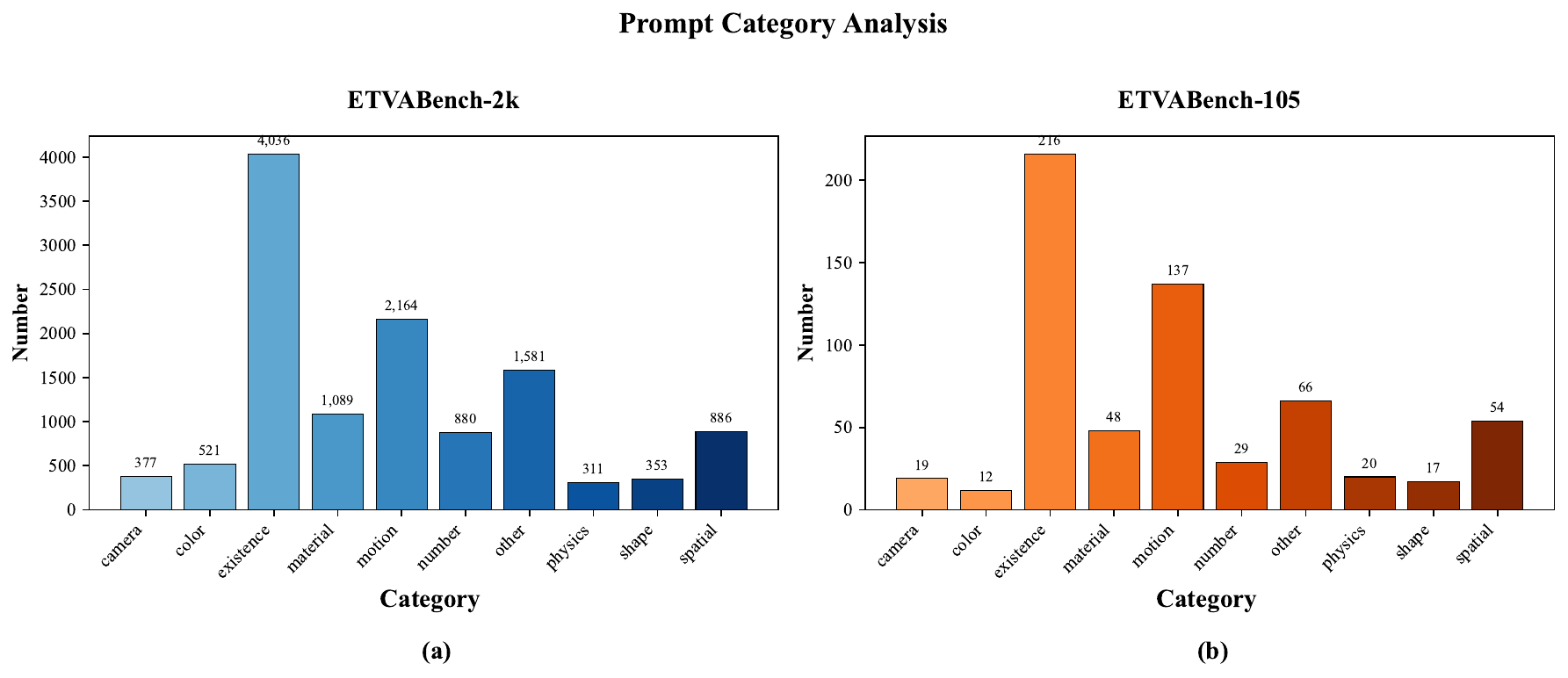}
    \vspace{-5mm}
    \caption{Prompt Category Distribution of ETVABench-2k and ETVABench-105}
    \label{fig:category_analysis}
    \vspace{-2mm}
\end{figure*}
\section{Details on Evaluation Categories}
The prompts were divided into 10 categories: \texttt{existence, action, material, spatial, number, shape, color, camera, physics, other}. Figure \ref{fig:category_analysis} shows that ETVABench-2k and ETVA-105 generate questions with similar distributions across these categories, indicating consistent coverage. Table \ref{tab:prompts_example} shows question example and prompt example for each category.

\begin{table*}[t]
    \centering
    \resizebox{\linewidth}{!}{
    \begin{tabular}{c|cc}
    \toprule
     Category &  Question Example & Prompt Example \\
     \midrule
    Existence & \textit{Is there a penguin in the video?} & \textit{A penguin standing on the left side of a cactus in a desert.}\\
    Action & \textit{Does the player pass the football?}& \textit{In a crucial game moment, a player passes the football, dodging opponents.}\\
    Material & \textit{Is the city made of crystals?} & \textit{A city made entirely of glowing crystals that change colors based on emotions.}\\
    Spatial & \textit{Does the penguin stand on the left of the cactus?} & \textit{A penguin standing on the left side of a cactus in a desert.}\\
    Number & \textit{Are there three owls in the video?} &\textit{Three owls perch on a branch during a full moon.} \\
    Shape & \textit{Is the cloud shaped like a hand?} & \textit{A cloud shaped like a giant hand that picks up people for transportation.}\\
    Color & \textit{Does the man's hair brown?} & \textit{There's a person, likely in their mid-twenties, with short brown hair.}\\
    Camera & \textit{Is the camera pushing in?} & \textit{A girl is walking forward, camera push in.}\\
    Physics & \textit{Is the water pouring out in the space station?} & \textit{Water is slowly pouring out of glass cup in the space station.}\\
    Other & \textit{Is the video in the Van Gogh style?} & \textit{A beautiful coastal beach waves lapping on sand, Van Gogh style.}\\
    \bottomrule
    \end{tabular}%
    }
    \vspace{-2mm}
    \caption{Question Example and Prompt Example for Each Category.}
    \label{tab:prompts_example}
\end{table*}
\section{Detailed results on ETVABench-2k}
More details of 10 open-source T2V models is in Table \ref{tab:full}.
\begin{table*}[h]
    \centering
\begin{adjustbox}{width=\linewidth,center}
\renewcommand{\arraystretch}{1}
\setlength{\tabcolsep}{1.2mm}
\begin{tabular}{lcccccccccc|c}
\toprule  
Model & Existence & Motion & Material & Spatial & Number & Shape & Color & Camera & Physics & Other & Avg \\
\midrule
\rowcolor{gray!20} \textit{Open-Source T2V Models} & & & & & & & & & & &  \\
Latte~ & 0.505 & 0.504 & 0.538 & 0.558 & 0.495 & 0.567 & 0.563 & 0.536 & 0.490 & 0.529 & 0.519 \\
OpenSora-plan-1.1~ & 0.534 & 0.526 & 0.496 & 0.568 & 0.522 & 0.504 & 0.522 & 0.488 & 0.397 & 0.568 & 0.529 \\
OpenSora-plan-1.2~ & 0.590 & 0.585 & 0.576 & 0.635 & 0.590 & 0.611 & 0.595 & 0.582 & 0.545 & 0.610 & 0.601 \\
OpenSora-1.2~ & 0.666 & 0.685 & 0.626 & 0.688 & 0.637 & 0.655 & 0.623 & 0.666 & 0.526 & 0.667 & 0.660 \\
Cogvideox-2B~ & 0.679 & 0.670 & 0.615 & 0.713 & 0.658 & 0.655 & 0.627 & 0.629 & 0.600 & 0.691 & 0.668 \\
Vchitect-2.0~ & 0.686 & 0.691 & 0.635 & 0.731 & 0.681 & 0.712 & 0.652 & 0.668 & 0.545 & 0.700 & 0.682 \\
CogvideoX-5B~ & 0.700 & 0.700 & 0.648 & 0.769 & 0.710 & 0.670 & 0.664 & 0.671 & 0.584 & 0.698 & 0.694 \\
CogvideoX-1.5-5B~ & 0.704 & 0.714 & 0.686 & 0.716 & 0.722 & 0.670 & 0.637 & 0.674 & 0.606 & 0.727 & 0.702 \\
Mochi-1-preview~ & 0.736 & 0.735 & 0.666 & 0.771 & 0.705 & 0.726 & 0.656 & 0.668 & 0.548 & 0.745 & 0.720 \\
Hunyuan~ & \textbf{0.753} & \textbf{0.756} & \textbf{0.699} & \textbf{0.810} & \textbf{0.747} & \textbf{0.761} & \textbf{0.726} & \textbf{0.695} & \textbf{0.632} & \textbf{0.762} & \textbf{0.748} \\
\bottomrule
\end{tabular}
\end{adjustbox}
\vspace{2mm}
\caption{ETVABench-2k evaluation results with 10 open-source T2V models and apple\_video. A higher score indicates better performance for a dimension. \textbf{Bold} stands for the best score.}
\label{tab:full}
\end{table*}
\section{Details of Text-to-Video Models}
\paragraph{Sora} Sora~\cite{sora2024} is a state-of-the-art text-to-video model built upon the DiT~\cite{Peebles2022DiT} architecture. As one of the most advanced and widely discussed video generation models, Sora is developed by OpenAI, though many of its underlying details remain undisclosed.  Notably, OpenAI has not released an API for Sora, restricting video generation to browser-based access, which is both resource-intensive and inconvenient. In this study, we adopt a 16:9 aspect ratio at 480p resolution to generate 5-second video samples.
\paragraph{Vidu} Vidu~\cite{vidu}, a text-to-video (T2V) diffusion model developed by Shengshu Technology, integrates advanced semantic comprehension with dynamic shot composition capabilities to achieve hierarchical video synthesis across resolutions ranging from low-definition to 1080p. Our experimental framework employed Vidu 1.5 to generate 4-second video sequences at 720p resolution (16:9 aspect ratio), quantitatively assessing its dual capacity for contextual fidelity and cinematographic control. 
\paragraph{Pika} Pika~\cite{pika}, a proprietary video synthesis model developed by Pika Labs, demonstrates versatile capabilities in both video generation and multimodal editing across diverse visual styles. For experimental validation, we utilized the Pika 1.5 implementation to synthesize 5-second video sequences at a 16:9 aspect ratio, systematically evaluating its capacity to preserve temporal consistency and stylistic fidelity under standardized conditions. 
\paragraph{Kling} Kling~\cite{kling} is a series of proprietary video generation models developed by Kuaishou.It is built on the Diffusion Transformer (DiT) architecture and has demonstrated exceptional capabilities in generative tasks.  For our experiments, we utilized Kling-1.0 and Kling-1.5, both standard models with a 16:9 aspect ratio, to generate 5-second video sequences. 
\paragraph{Hunyuan-Video} Hunyuan-Video~\cite{kong2024hunyuanvideo}, developed by Tencent, is currently the most advanced open-source T2V model. It features a massive 13 billion parameters architecture and is trained using the flow matching~\cite{lipman2022flow} method on a hierarchically structured, high-fidelity video dataset. For our experiments, we set the resolution to 544×966, generating a 5-second video consisting of 121 frames with 24 fps.
\paragraph{Mochi}Mochi~\cite{genmo2024mochi}, a text-to-video (T2V) diffusion model developed by Genmo, demonstrates significant advancements in video synthesis through its 10-billion-parameter architecture. Preliminary evaluations indicate exceptional performance in motion fidelity and textual prompt alignment, substantially reducing the quality disparity between proprietary and open-source video generation systems. For experimental validation, we generated a 2-second video sequence (61 frames at 24 fps) with 480×848 pixel resolution, effectively demonstrating the model's temporal coherence and detail preservation capabilities.
\paragraph{CogVideox} CogVideoX~\cite{yang2024cogvideox} is a large-scale open-source T2V model released by Zhipu, available in three versions: CogVideoX-2B, CogVideoX-5B, and CogVideoX-1.5-5B. It incorporates a 3D causal VAE and an expert transformer, enabling the generation of coherent, long-duration, and high-action videos. For CogVideoX-2B and CogVideoX-5B, we set the frame rate to 8 fps, generating a 6-second video with a total of 49 frames at a resolution of 720×480. For CogVideoX-1.5-5B, we used a resolution of 1360×768 at 16 fps, producing a 5-second video with 91 frames.
\paragraph{Opensora} OpenSora~\cite{opensora} is a high-quality DiT-based text-to-video model that introduces the ST-DiT-2 architecture, It supports flexible video generation with varying aspect ratios, resolutions, and durations. The model is trained on a combination of images and videos collected from open-source websites, along with a labeled self-built dataset. We utilized the officially released OpenSora 1.2 code and model, setting the spatial resolution to 720p and the frame rate to 24 fps, producing a 4-second (96-frame) video.
\paragraph{Opensora-plan} OpenSoraPlan~\cite{lin2024opensora-plan} is an advanced video generation model built upon Latte~\cite{ma2024latte}. It replaces the Image VAE~\cite{kingma2022vae} with Video VAE~\cite{lin2024opensora-plan} (CausalVideoVAE), following a similar approach to Sora~\cite{sora2024}. For OpenSoraPlan v1.1, we used the 65-frame version with a spatial resolution of 512×512 at 16 fps, generating a 4-second video. For OpenSoraPlan v1.2, we selected the 93-frame version with a resolution of 720p at 16 fps, producing a 5-second video.
\paragraph{Vchitect} Vchitect~\cite{fan2025vchitect}, an open-source text-to-video (T2V) generative model developed by Shanghai AI Lab, is built upon the Diffusion Transformer (DiT)~\cite{Peebles2022DiT} architecture. With 2 billion parameters, the model demonstrates robust capabilities in generating high-quality video content. For experimental validation, we synthesized a 5-second video sequence comprising 40 frames at 8 frames per second (fps), with a resolution of 768×432 pixels. 
\paragraph{Latte} Latte~\cite{ma2024latte} is an early open-source DiT-based text-to-video model, built upon PixArt-Alpha with extended spatiotemporal modules and further training. We employed the officially released LatteT2V code and model, preserving the original parameter settings. For video generation, we used a spatial resolution of 512×512, a frame rate of 8 fps, and a duration of 2 seconds (16 frames).
\section{Details of Human annotation}
\begin{figure*}[!th]
    \includegraphics[width=\linewidth]{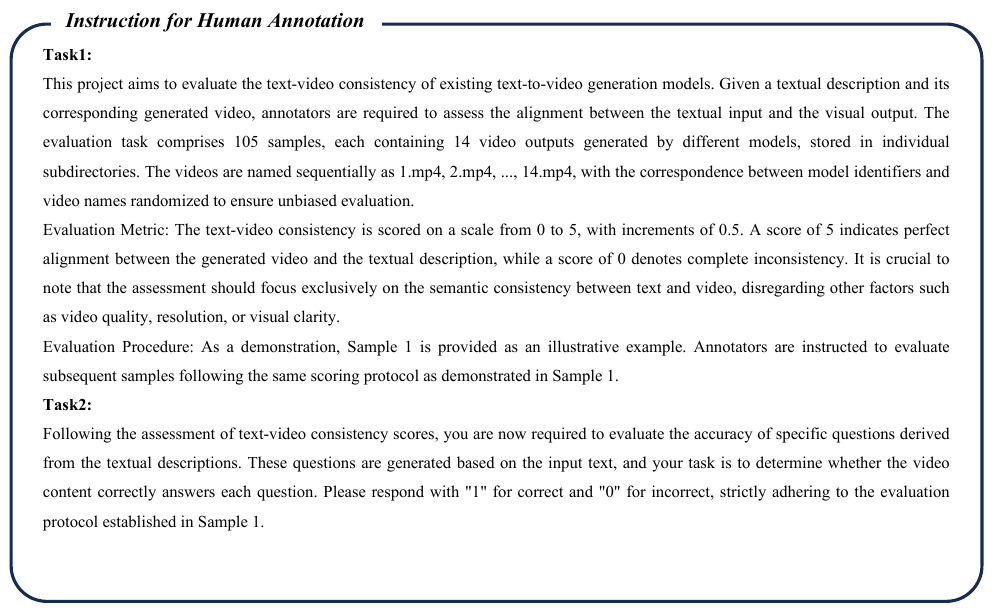}
    \caption{Instruction for Human Annotation}
    \label{fig:prompt1}
    \vspace{-2mm}
\end{figure*}

Figure \ref{fig:prompt1} shows detailed instructions for human annotation.
\section{Prompts of ETVA}
\begin{figure*}[!th]
    \includegraphics[width=\linewidth]{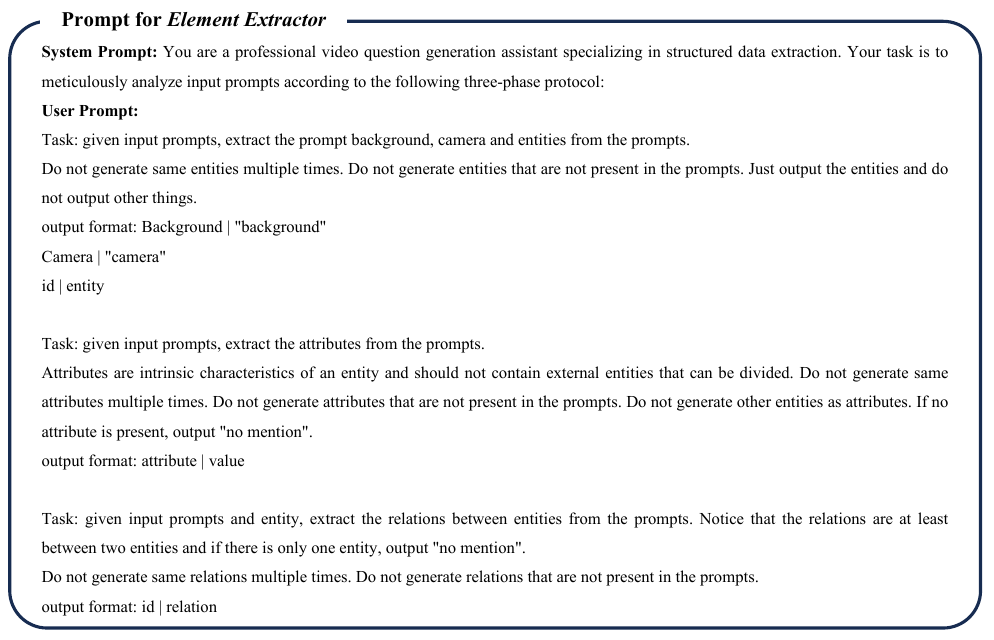}
    \caption{Prompt for \textit{Element Extractor}.}
    \label{fig:prompt2}
    \vspace{-2mm}
\end{figure*}

\begin{figure*}[!th]
    \includegraphics[width=\linewidth]{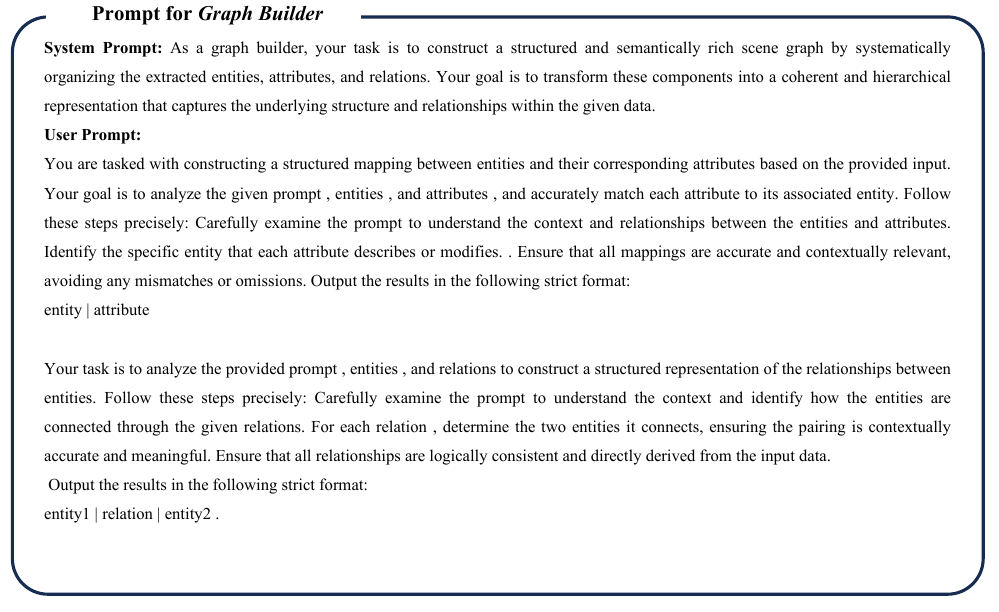}
    \caption{Prompt for \textit{Graph Builder}.}
    \label{fig:prompt3}
    \vspace{-2mm}
\end{figure*}

\begin{figure*}[!th]
    \includegraphics[width=\linewidth]{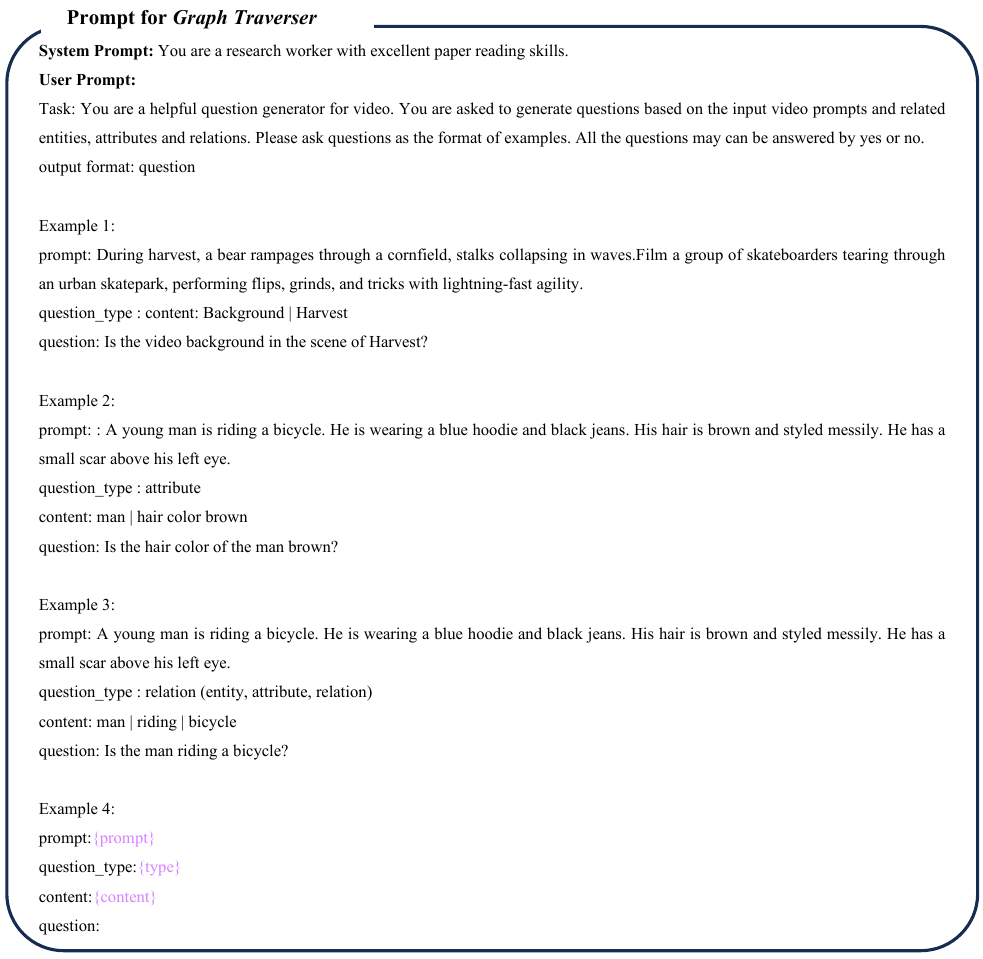}
    \caption{Prompt for \textit{Graph Traverser}.}
    \label{fig:prompt4}
    \vspace{-2mm}
\end{figure*}

\begin{figure*}[!th]
    \includegraphics[width=\linewidth]{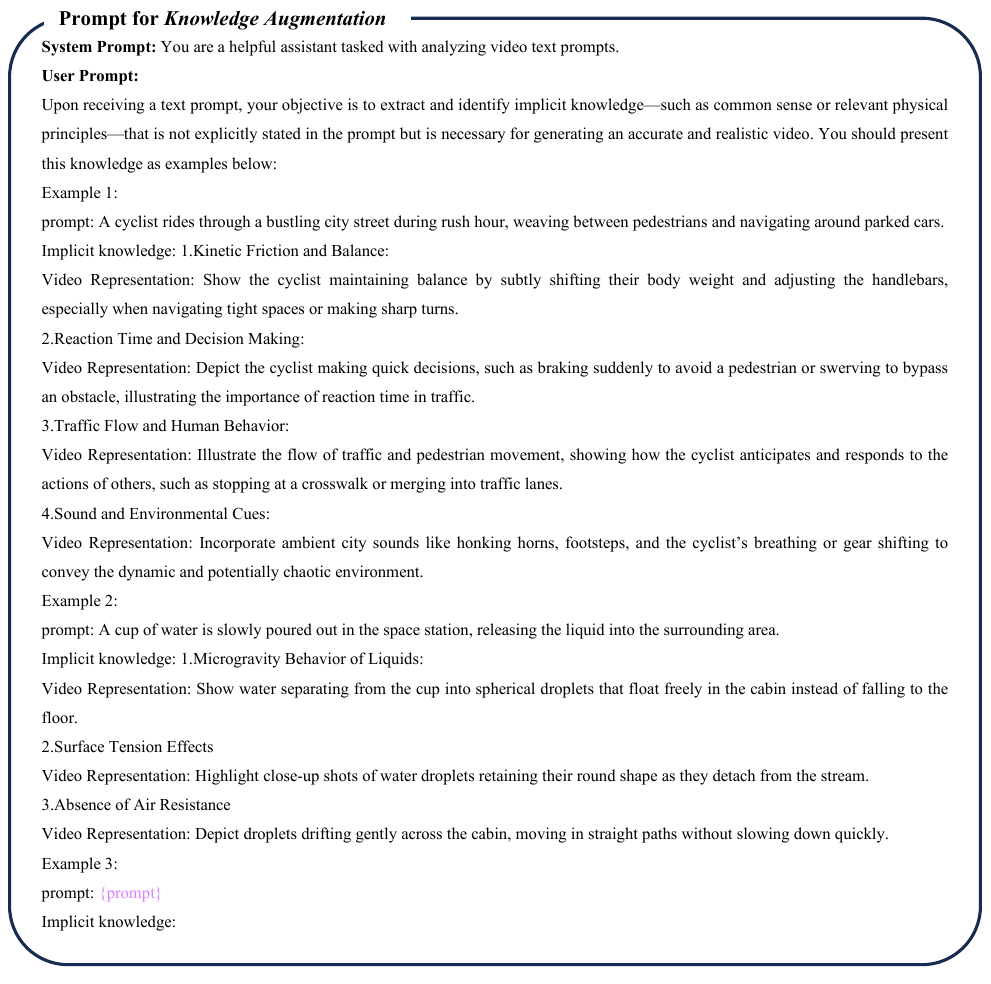}
    \caption{Prompt for \textit{Knowledge Augmentation}.}
    \label{fig:prompt5}
    \vspace{-2mm}
\end{figure*}

\begin{figure*}[!th]
    \includegraphics[width=\linewidth]{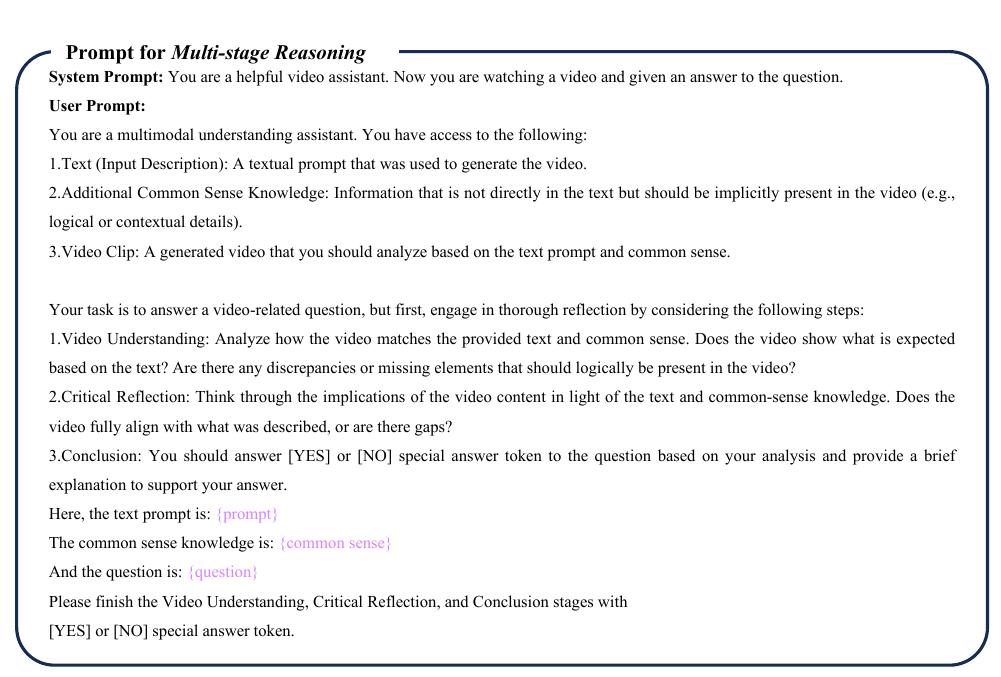}
    \caption{Prompt for \textit{Multi-stage Reasoning}.}
    \label{fig:prompt6}
    \vspace{-2mm}
\end{figure*}

Figure \ref{fig:prompt2} - Figure \ref{fig:prompt6} present the detailed prompts in ETVA.
\section{More Cases about ETVA}
\begin{figure*}[!ht]
    \centering
    \includegraphics[width=\linewidth]{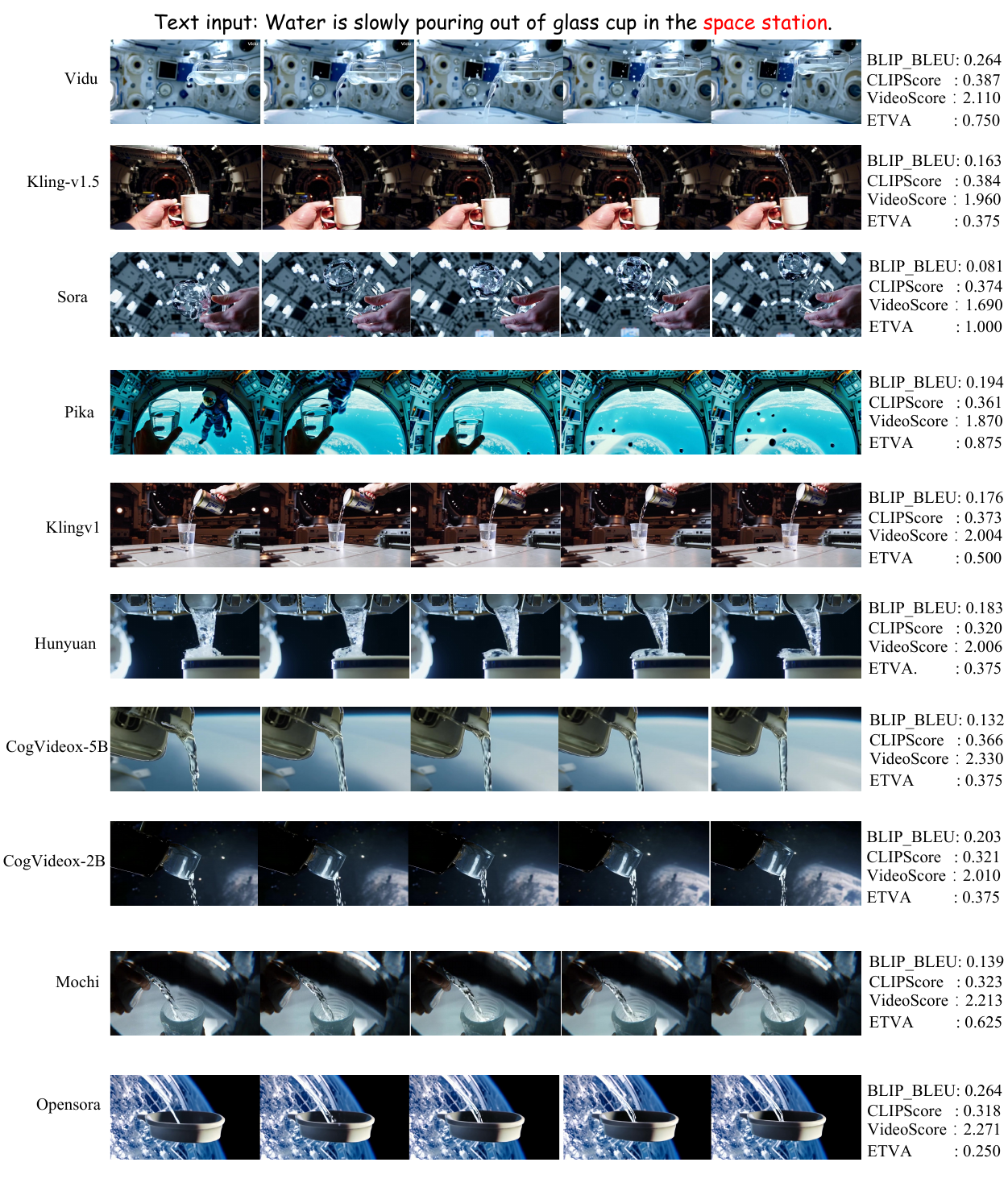}
    \caption{ Illustration of a comparative analysis between ETVA and conventional evaluation metrics, based on the text prompt: ``Water is slowly pouring out of glass cup in the space station". We compare our ETVA score with conventional text-to-video alignment metrics.}
    \label{fig:case1}
    \vspace{-2mm}
\end{figure*}

\begin{figure*}[!ht]
    \centering
    \includegraphics[width=\linewidth]{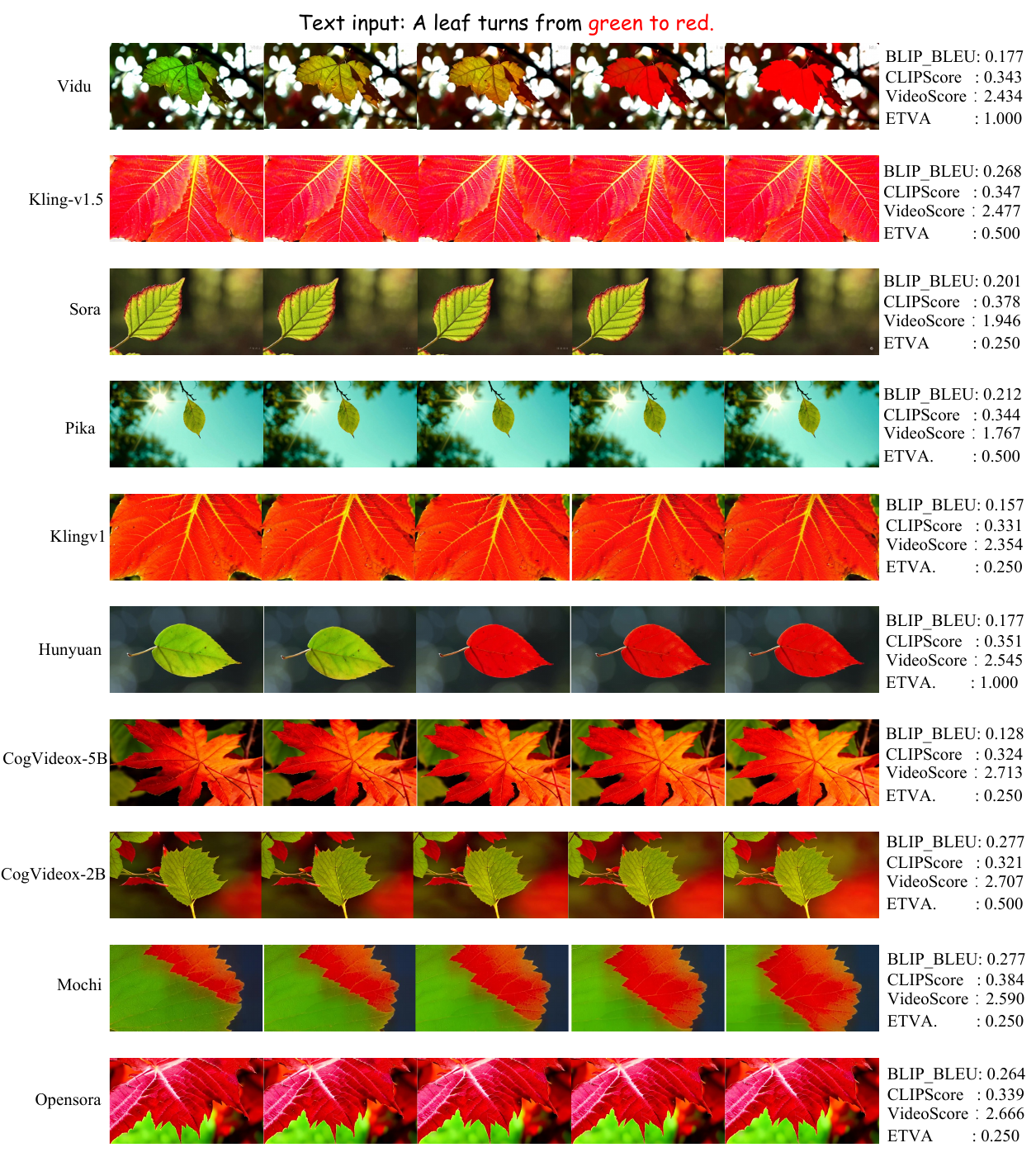}
    \caption{Illustration of a comparative analysis between ETVA and conventional evaluation metrics, based on the text prompt: "A leaf turns from green to red". We compare our ETVA score with conventional text-to-video alignment metrics.}
    \label{fig:case2}
    \vspace{-2mm}
\end{figure*}

Figure \ref{fig:case1} and Figure \ref{fig:case2} show the detailed comparision results  between ETVA and conventional evaliation metrics.
\clearpage

\clearpage




\end{document}